\documentclass[12pt]{elsarticle}

\makeatletter
\def\ps@pprintTitle{%
 \let\@oddhead\@empty
 \let\@evenhead\@empty
 \def\@oddfoot{}%
 \let\@evenfoot\@oddfoot}
\makeatother

\usepackage{amssymb}

\RequirePackage{amsmath}
\usepackage{longtable}
\usepackage{booktabs}
\usepackage{array}
\usepackage[ruled]{algorithm2e}
\usepackage{multirow}
\usepackage[ruled]{algorithm2e}
\usepackage{theorem}

\newtheorem{theorem1}{Special Theorem}
\newtheorem{definition}[theorem1]{Definition}

\newtheorem{theorem}[theorem1]{Theorem}

\newtheorem{lemma}[theorem1]{Lemma}

\graphicspath{ {images/} }
\begin{document}

\begin{frontmatter}



\title{Efficient Clustering of Correlated Variables and Variable Selection in High-Dimensional Linear Models}

 

\fntext[fn1]{Indian Statistical Institute, Bangalore Centre}
\fntext[fn2]{Indian Statistical Institute, Kolkata}

\author{Niharika Gauraha\fnref{fn1}}
\author{and Swapan K. Parui\fnref{fn2}}
\address{Indian Statistical Institute}

\begin{abstract}
In this paper, 
we introduce Adaptive Cluster Lasso(ACL) method for variable selection in high dimensional sparse regression models with strongly correlated variables. To handle correlated variables, the concept of clustering or grouping variables and then pursuing model fitting is widely accepted. When the dimension is very high, finding an appropriate group structure is as difficult as the original problem. 
The ACL is a three-stage procedure where, at the first stage, we use the Lasso(or its adaptive or thresholded version) to 
do initial selection, then we also include those variables which are not selected by the Lasso but are strongly correlated with the variables selected by the Lasso. 
At the second stage we cluster the variables based on the reduced set of predictors and in the third stage we perform sparse estimation such as Lasso on cluster representatives or the group Lasso based on the structures generated by clustering procedure. We show that our procedure is consistent and efficient in finding true underlying population group structure(under assumption of irrepresentable and beta-min conditions). We also study the group selection consistency of our method and we support the theory using simulated and pseudo-real dataset examples.
\end{abstract}


\end{frontmatter}


\section{Introduction}
We consider the usual linear regression model 
\begin{align} \label{eq:lr}
	\textbf{Y} &=  \textbf{X} \beta^{0}+\epsilon,
\end{align}
with response vector $Y_{n \times 1}$, design matrix $X_{ n \times p}$, true underlying coefficient vector $\beta^0_{p \times 1}$ and error vector $\epsilon_{n\times 1}$.  
When the number of predictors (p) is much larger than the number of observations (n), $p >> n$, the ordinary least squares estimator is not unique and may over fit the data. The parameter vector
$\beta^{0}$ can only be estimated based on given very few observations under assumption of sparsity in $\beta^{0}$.  To infer the true active set $S_0 = \{j; \beta^{0}_{j} \neq 0\}$, the Lasso(\cite{Tibshirani}), its variants and other regularized regression methods are mostly used for sparse estimation and variable selection. However, variable selection in situations involving high empirical correlation remains one of the most important issues. This problem is encountered in many applications such as in microarray analysis, a group of genes which participate in the same biological pathway tend to have highly correlated expression levels(see \cite{Segal}) and it is often required to consider all of them if they are related to the underlying biological process.

It has been proven that the design matrix must satisfy the following conditions for the Lasso to perform exact variable selection: irrepresentable(IR) condition(\cite{Zhao}) and beta-min condition(\cite{Buhlmann1}). Having highly correlated variables implies that the design matrix violates the IR condition. To deal with variable selection with correlated variables, mainly two approaches has been proposed: simultaneous clustering and model fitting and clustering followed by the sparse estimation (i.e. group lasso). The former approach imposes restrictive conditions on the design matrix. However, the time complexity for clustering of variables severely limits the dimension of data sets that can be processed by the later approach. Moreover, group selection in models with a larger number of groups is more difficult(see \cite{Wei}). 
To overcome these limitations we propose a three stage procedure, Adaptive Cluster Lasso method. 
Basically, we try to reduce the dimensions first using the Lasso(or its adaptive or thresholded version) before clustering of variables. 

At a high level our method works as follows. At the first stage, the Lasso is used to do initial selection of variables 
then we also select those variables which are not selected by the Lasso but they are strongly correlated with the variables selected by the Lasso. If the design matrix satisfies the beta-min condition, then after the first stage, the selected set of variables contains the true active set and the dimensionality of the problem is reduced by a huge amount. At the second stage, we perform clustering of variables on the reduced model, so that strongly correlated variables are grouped together in disjoint clusters. In the third stage, we do group-wise sparse estimation based on the structures generated by clustering procedure. The second and third stages of ACL together is the same as Cluster Group Lasso(CGL) or Cluster Representative Lasso(CRL) with ordinary hierarchical correlation based clustering, defined in \cite{Buhlmann2}.  Hence, ACL method is an extension of the clustering lasso methods proposed in \cite{Buhlmann2}.

Mainly, there are two lines of thought, the one is to find an appropriate and efficient clustering of correlated variables and the other line of thought is to avoid the false negatives. 
With these thoughts in mind, we develop a computationally efficient variable selection procedure $\hat{S}_{ACL}$, which identifies the appropriate correlated group structures and selects all variables from a group of correlated variables where at least one of them is active. Assuming Group Irrepresentable(GIR) and group beta-min condition on the design matrix, we prove that the $\hat{S}_{ACL}$ selects the true model, with much less computational complexity. We show that the dimensionality reduction and subsequent clustering and CGL(or CRL) improves over the plain clustering and CGL(or CRL). 
We illustrate the proposed method and compare it with the methods proposed in \cite{Buhlmann2} by extensive simulation studies and we also apply it to a pseudo-real dataset. 

The rest of this paper is organized as follows. In section 2, we provide notations, assumptions, review of relevant work and we discuss our contribution. In section 3, we review mathematical theory of the Lasso and group Lasso. In section 4, we describe the proposed algorithm which mostly selects more adequate models in terms of model interpretation and prediction performance. In section 5, we study theoretical properties of the proposed method. We also show that the variable selection is consistent for high dimensional sparse problems. In section 6, we provide numerical results based on simulated and pseudo real dataset. Section 7 contains the computational details and we shall provide conclusion in section 8. 

\section{Background and Notations}
In this section, we state notations and assumptions, we define required concepts and we also provide review of the relevant work.

\subsection{Notations and Assumptions}
The following notations,assumptions and definitions are applied throughout this paper. 

We consider the usual linear regression set up with univariate response variable $Y \in R$ and p-dimensional predictors $X_i \in R^p$:
\begin{align} \label{eq:lr2}
	Y_i& =  \sum_{j=1}^{p} X_{i}^{(j)}\beta^{0}_{j} + \epsilon_i \quad i=1,...,n \; \;j=1,...,p
\end{align} 
or, in matrix notation (as in Equation \ref{eq:lr})
\begin{align*}
	\textbf{Y} &=  \textbf{X} \beta^{0}+\epsilon
\end{align*}
where $\beta^{0} \in R^p$ are unknown true regression coefficients to be estimated, and the components of the noise vector $\epsilon \in R^n$ are i.i.d. $N(0, \sigma^2)$. The columns of the design matrix X are denoted by $X^{j}$. We assume that the design matrix \textbf{X} is fixed, the data is centred and the predictors are standardized, so that 
$\sum_{i=1}^{n} Y_i = 0$, $\sum_{i=1}^{n} X^{j}_{i} = 0$ and $\frac{1}{n} X^{j'}X^{j} = 1$ for all $j=1,...,p$.\\
The $L_1$-norm is defined as:
\begin{align} \label{eq:l1}
	\|\beta\|_1 = \textstyle \sum_{j=1}^p |\beta_j|
\end{align}
$L_2$-norm squared is defined as:
\begin{align} \label{eq:l2}
	\|\beta\|^{2}_2 = \textstyle \sum_{j=1}^p \beta^{2}_{j}
\end{align}
The $L_\infty-$ norm is defined as:
\begin{align} \label{eq:linf}
	\|\beta\|_{\infty} = \textstyle max_{1 \leq i \leq n |} |\beta_j|
\end{align}
The true active set  $S_0$ denotes the support of the subset selection solution($S_0 = supp(\beta_0)$) and defined as
\begin{align} \label{eq:s0}
	S_0 &= \{j; \beta^{0}_{j} \neq 0\}
\end{align}
The sign function is defined as:
\begin{align}
	sign(x) = \left\lbrace \begin{array}{ll}
	-1 & \text{ if } x < 0  \\
	0 & \text{ if }  x = 0\\
	1 & \text{ if }  x > 0
\end{array}	 \right.
\end{align}
The (scaled)Gram matrix(covariance matrix) is defined as
\[
	\hat{\Sigma}= \frac{X'X}{n}
\]
The $\beta_S$ has zeroes outside the set S,as
\[
	\beta_S = \{ \beta_j I(j \in S) \}
\]
and $\beta = \beta_S + \beta_{S^c}$.\\
For the given $S \subset \{1,2,...,p  \}$, the covariance matrix can be partitioned as:
\begin{align} \label{sigmaPart}
	\Sigma = \left[ \begin{array}{cc}
	\Sigma_{11} = \Sigma(S) & \Sigma_{12}(S)\\
	\Sigma_{21}(S)\quad \quad &\Sigma_{22} = \Sigma(S^c)	
	\end{array} \right]
\end{align}
Minimum eigenvalue of a matrix A is denotes as $\Lambda_{min}(A)$.

\subsection{Clustering of Variables} \label{sub:clust}
We use correlation based, bottom-up agglomerative hierarchical clustering methods to cluster predictors, which forms groups of variables based on correlations between them. For further details on grouping of variables and determining the number of clusters, we refer to \cite{Buhlmann2}. 

\subsection{The Lasso and the Group Lasso}
The Least Absolute Shrinkage and Selection Operator (Lasso) was introduced by Tibshirani \cite{Tibshirani}. It is a penalized least squares method that imposes an L1-penalty on the regression coefficients, which does both shrinkage and automatic variable selection simultaneously due to the nature of the L1-penalty. 

We denote $\hat{\beta}$, as a Lasso estimated parameter vector. Assume $\lambda$ is the regularization parameter, then then Lasso estimator is computed as:
\begin{align}
	\hat{\beta} \in \mathop{arg min}_{\beta \in \mathbb{R}^p} \{\frac{1}{n} \| \textbf{y} - \textbf{X} \beta \|_{2}^{2}+ \lambda \|\beta\|_1 \}
\end{align}
and the estimated active set is denoted as $\hat{S}$ and defined as
\begin{align} \label{eq:s1}
	\hat{S} = \{j; \hat{\beta}_{j} \neq 0\}
\end{align}

The Lasso error vector is defined as 
\begin{align}
	\Delta = \hat{\beta} - \beta^0
\end{align}

One of the major disadvantages of the the lasso is that, the Lasso tends to select single or a few variables, from a group of highly correlated variables. When the distinct groups or clusters among the variables are known a priory and it is desirable to select or drop the whole group instead of single variables. Then the Group Lasso (see \cite{Yuan}) or its variants are used, that imposes an $L_2$-penalty on the coefficients within each 
group to achieve such group sparsity. 

Here we define some more notations and state assumptions for the group Lasso. We may interchangeably use $\beta^{0}$ and $\beta$ for the true regression coefficient vector, the later one is without the superscript. Let us assume that the parameter vector $\beta$ is structured into groups, $G = \{ G_1, . . . , G_q \}$, where $q < n $, denotes the number of groups. The partition $G$ basically builds a partition of the index set $\{ 1,...,p\}$ with $ \cup_{r=1}^{q} G_r = \{1,... , p\}$ and $G_r \cap G_l = \emptyset, \quad r\neq l$. The parameter vector $\beta$, then has the structure $\beta = \{ \beta_{G_1}, ..., \beta_{G_q} \}$ where $\beta_{G_j} = \{\beta_r: r \in G_j \}$. 

The columns of the each group is represented by $X^{G_j}$.
\[
	X = (X^{(1)}, ..., X^{(p)}) = (X^{(G_1)}, ... , X^{(G_q)})
\]
The response vector Y can also be written as
\[
	Y = \sum_{j=1}^{q} X^{(G_j)} \beta_{G_j} + \epsilon
\]
where $X^{(G_j)} \beta_{G_j} = \sum_{k=1}^{m_k} X^{(G_j)}_k (\beta_{G_j})_k $.
The loss function of the group Lasso is same as the loss function of the Lasso $\frac{1}{n}\|Y-X\beta \|_{2}^{2}$. The group Lasso penalty is defined as
\[
	\|\beta\|_{2,1} = \textstyle \sum_{j=1}^q \| X^{G_j} \beta_{G_j} \|_2 \sqrt{\frac{m_j}{n}}
\]
where $m_j = |G_j|$ is the group size. 
Since the penalty is invariant under parametrizations within-group. Therefore, without loss of generality, we can assume $\Sigma_{rr} = I$, the $m_r \times m_r$ identity matrix. Hence the group Lasso penalty can be written as
\[
	\|\beta\|_{2,1} = \sum_{j=1}^q \sqrt{m_j} \|\beta_{G_j} \|_2
\]

The Group Lasso estimator(with known q groups) is defined as
\begin{align} \label{eq:grpLasso}
	\hat{\beta}_{grp}\in \mathop{argmin}_{\beta} \{ \frac{1}{n}\|Y-X\beta \|_{2}^{2} + \lambda \|\beta\|_{2,1} \}
\end{align}


The group Lasso has the following properties:
\begin{itemize}
\item The group Lasso behaves like the lasso at the group level,
depending on the value of the regularization parameter $\lambda$, the whole group of variables may drop out of the model.
\item For singleton groups (when the group sizes are all one), it reduces exactly to the lasso.
\item The group Lasso penalty is invariant under orthonormal transformation within the groups.
\item The group Lasso estimator has similar oracle inequalities as the standard Lasso for prediction accuracy and estimation error. It has group wise variable selection property(We discuss mathematical theory in the section \ref{secLassoTheory}).
\end{itemize}
   
Let W denote the actives group set , $W \subset \{ 1,...,q \}$, with cardinality $w = |W|$. Throughout the article, the following assumption are made for the group Lasso:
\begin{itemize}
 \item[$(A1)$:] The size of the each group is less than the number of observations.
	\[ m_{max} < n \].
\item[$(A2)$:]  The number of active groups, w, is less than the number of observations (sparsity assumption). 
\end{itemize}

\subsubsection{Cluster Group Lasso}
When the group structure is not known then clusters $G_1, . . . , G_q$ are generated from the design matrix X( using correlation based method etc.). Then the group Lasso is applied to the resulting clusters. We denote the clusters selected by the group Lasso as $\hat{S}_{clust}$, and is defined as
\begin{align}
		\hat{S}_{clust} = \{r: \text{ cluster }G_r \text{ is selected, } r = 1,...,q\}
\end{align}
The union of the selected clusters gives the selected set of variables.
\begin{align}
	\hat{S}_{CGL} = \cup_{r \in \hat{S}_{clust}} G_r
\end{align}
\subsubsection{Cluster Representative Lasso}
Similar to the CGL, the cluster representative Lasso, first identifies groups among the variables and then applies the lasso for cluster representatives (see \cite{Buhlmann2}). When sign of the regression coefficients within a group is the same then taking group representatives  is advantageous, whereas when near cancellation among $\beta^{0}_{j}(j \in G_r)$ takes place then CGL is preferred.
\\
We define representative for each cluster as
\begin{align*}
\bar{X}^{(r)} = \frac{1}{|G_r|} \sum_{j \in G_r} X^{(j)}, \quad r = 1,...,q.
\end{align*}
The design matrix of cluster representatives is denoted as  $\bar{X}_{n \times q}$.
Then optimization problem for CRL is defined using response $Y$ and the design matrix of cluster representatives $\bar{X}$ as:
\begin{align}
	\hat{\beta}_{CRL} \in \arg\!\min_{\beta} ( \|{ \textbf{y} - \bar{\textbf{X}} \beta} \|^{2}_{2}+ \lambda_{CRL} \|\beta \|_1 )
\end{align}

The selected clusters are then denoted as:
\begin{align*}
	\hat{S}_{clust,CRL} = \{r; \hat{\beta}_{CRL,r} \neq 0, r = 1, . . . , q \}
\end{align*}
 
and the selected variables are obtained as the union of the selected clusters as:
\begin{align*}
	\hat{S}_{CRL} = \cup_{r \in \hat{S}_{clust,CRL}} G_r
\end{align*}

\subsection{Review of Relevant work and our Contribution}
Here, we provide a brief review of relevant work in this area, and we also show that how our proposal differs or extends the previous studies.

The Lasso can not do variable selection in the situations where predictors are highly correlated. As mentioned before, to handle correlated covariates in variable selection methods, two algorithmic approaches have been developed in the past: clustering of variables and model fitting either simultaneously or at two different stages. 
Examples of the methods that do clustering and model fitting simultaneously are Elastic Net(\cite{Hui}), Fused LASSO(\cite{Fused}), octagonal shrinkage and clustering algorithm for regression(OSCAR, \cite{oscar}) and Mnet(\cite{Mnet} ) etc. The Elastic Net uses a combination of the $L_1$ and $L_2$ penalties, OSCAR uses a combination of  $L_1$ norm and and $L_{\infty}$ norm and Mnet uses a combination of $L_2$ and Minimum Concave Penalty(MCP). We note that these methods use only combination of penalties, they do not use any specific information on the correlation pattern among the predictors and hence they do not reveal any group structure in the data. 

Now, we discuss a few methods that perform clustering and model fitting at different stages, i.e. 
Cluster Group Lasso(CGL, \cite{Buhlmann2}), Cluster representative Lasso(CRL,\cite{Buhlmann2}), Stability Feature Selection using Cluster Representative LASSO (SCRL, \cite{Niharika}) and sparse Laplacian shrinkage estimator(SLS, \cite{SLS}).
CRL, CGL and SCRL use correlation based and canonical correlation methods to perform hierarchical clustering. SLS also considers the correlation patterns among predictors but requires that highly correlated variables should have similar predictive effects.
The main disadvantage of this approach 
is mainly due to clustering in the presence of unstructured data or noise features. It is difficult to determine the exact group structures or the exact number of groups in high-dimensions and in the presence of noise features. Moreover, the CPU time taken by clustering algorithms is unacceptable when the number of predictors are huge. To address these problems, we propose to reduce the dimensionality before performing clustering which makes our proposal different from previous work. 

Basically, our work can be viewed as an extension of the two stage procedure, Cluster Lasso Methods with correlation based clustering, proposed in \cite{Buhlmann2}. The extension is that we add a dimensionality reduction stage prior to performing the clustering, which leads to clustering of variables more accurately and efficiently and thus consistent group variable selection. In particular, we consider  Adaptive Clustering Group Lasso(ACGL), where the Lasso is used as preprocessing step at the first stage, correlation based clustering at second stage and the group Lasso in the third stage(defined in section 4). We also consider the CGL method with ordinary hierarchical clustering, denoted by CGLcor, see \cite{Buhlmann2}. We compare ACGL and CGLcor in terms of predictive performance, variable selection and CPU time expended, in section 5. Our extensive simulation studies show that ACGL outperforms the CGLcor. 


\section{Mathematical Theory of the Lasso and the Group Lasso} \label{secLassoTheory}
In this section, we review the results required for proving consistent variable selection( and group variable selection) in high dimensional linear models. For more details on the mathematical theory for the lasso and group lasso, we refer to: \cite{Sara2}, \cite{Zhao}, \cite{Wei} \cite{Buhlmann2} and \cite{Buhlmann1}.
\begin{definition}[Compatibility Condition]
The Lasso compatibility condition holds for a fixed set $S \subset \{1,...,p\}$ with cardinality $s = |S|$, a constant $\phi_{comp}(S) > 0$ and if for all $ \| \Delta_{S^c}\|_1 \leq 3 \| \Delta_{S}\|_1 \neq 0$ the following holds
\begin{align*}
	\| \Delta_{S}\|_{1}^{2} \leq  \left\lbrace \frac{s\frac{1}{n} \| X \Delta \|_{2}^{2} }{ \phi^{2}(S)}  \right\rbrace
\end{align*}
where  $\phi_{comp}(S) $ is called the compatibility constant.
\end{definition}
The constant 3 is due to the condition $\lambda \geq 2\lambda_0$, which is required to overrule the stochastic process part,(see \cite{Buhlmann1} for details). Without loss of generality we can assume $S = \{1,...,s \}$ and partition the covariance matrix in block-wise form as given in equation \ref{sigmaPart}. Assuming $\Sigma^{-1}_{11}$ is invertible, the various form of Irrepresentable(IR) Conditions are defined as follows.

\begin{definition}
The strong irrepresentable condition is said to be met for the set S, with cardinality $s = |S|$, if the following holds:
\begin{align} \label{IR1}
	\|\Sigma_{12}(S) \Sigma^{-1}(S) \tau_S  \|_{\infty} < 1, \quad \forall \tau_S \in \mathbb{R}^s \; such \; that \;   
	\| \tau_S \|_{\infty} \leq 1
\end{align}
The weak irrepresentable condition holds for a fixed $\tau_S \in \mathbb{R}^s$ if 
\begin{align} \label{IR2}
	\|\Sigma_{12}(S) \Sigma^{-1}(S) \tau_S  \|_{\infty} \leq 1
\end{align}
For some $0<\theta <1$, the $\theta$ uniform irrepresentable condition holds if
\begin{align}\label{IR3}
	\mathop{max}_{\| \tau_S \|_{\infty} \leq 1} \|\Sigma_{12}(S) \Sigma^{-1}(S) \tau_S  \|_{\infty} \leq \theta
\end{align}
\end{definition}

Sufficient conditions(eigenvalue and mutual incoherence) on design matrix to hold IR are discussed in \cite{Zhao} and \cite{Hastie}.

\begin{definition}[Beta-min Condition]
 the Beta Min Condition is met for the regression coefficient $\beta^0$, if 
 $\min |\beta^0| \geq \frac{4 \lambda s_0}{\phi^{2}(S)}$
\end{definition}

\begin{lemma}
Under the following assumptions the Lasso selects the true active set $S_0$ with high probability:
\begin{itemize}
\item[$(A3)$:] Irrepresentable Condition holds for $S_0$.
\item[$(A4)$:] beta-min condition holds for $\beta^0$. 
\end{itemize}
\end{lemma}

The following inequality shows the bounds for prediction error and estimation error of the Lasso estimator.( for derivation and proof we refer to \cite{Buhlmann1}).
 \begin{align}
 	 \frac{1}{n}\| X \Delta \|_{2}^{2} +  \lambda \|\Delta \|_1  & \leq \frac{4\lambda^2 s}{ \phi^{2}_{comp}}
 \end{align}
Our error analysis for the group Lasso is based on the pure active group and pure noise group assumptions, that is, \\
$(A5):$ all variables are active variables within an active group and no variables are active in a noise group. \\

We define the group Lasso error as $\Delta_{G_r} =  \beta_{G_r} - \beta^{0}_{G_r} $, and also assume the following.\\
$(A6)$: We assume that clustering process identifies the group structure correctly.

\begin{definition}[The Group Lasso Compatibility Condition]
The group Lasso compatibility condition holds for a fixed set $W \subset \{1,...,q\}$ with cardinality $w = |W|$, a constant $\phi_{grp}(W) > 0$ and if for all $ \sum_{r \in W^c}\|  \Delta_{G_r} \|_{2} \leq 3 \sum_{r \in W}\| \Delta_{G_r} \|_{2}  \neq 0$ the following holds
\begin{align*}
	(\sum_{r \in W}\| \Delta_{G_r} \|_{2})^2  \leq  \left\lbrace \frac{ w\frac{1}{n} \| X \Delta \|_{2}^{2} }{\phi_{grp}^{2}(W)}  \right\rbrace
\end{align*}
where  $\phi_{grp}(W) $ is called the group Lasso compatibility constant.
\end{definition}

The Lasso compatibility condition implies the group Lasso compatibility condition, it is explained by the following Lemma(See Lemma 8.2 of the book \cite{Buhlmann1}, for the proof).

\begin{lemma}
	Let $W \subset \{ 1,...,q \}$ be a group index set, say, $W = \{ 1, ..., w\}$ Consider the full index set corresponding to W:
	 \[ S = \{ (1,1), ..., (1,m1),..., (w,1),...,(w,m_w) \} = \{ 1,...,s \}\], where (i,j) denotes  jth member of ith group and $s = \sum^{w}_{j=1} m_j$. If compatible condition holds for S with compatiblility constant $\phi(S)$ then the compatibility condition holds for the $\phi_{grp}(W)$, and  $\phi_{grp}(W) \geq \phi(S)$
\end{lemma}

\begin{definition}[Group Irrepresentability Condition]
The group IR condition is met for the set W with a constant $0< \theta < 1$, if for all $\tau \in \mathbb{R}^s$ with $\|\tau \|_{2, \infty} = \mathop{max}_{1\leq r \leq q} \| \tau_{G_r} \|_2 \leq 1 $, the following holds
\begin{align}
\frac{1}{m_r} \|(\Sigma_{21} \Sigma^{-1}_{11} K \tau)_{G_r} \| \quad\forall r \not\in W,
\end{align}
where $K = diag(m_1 I_{m_1} , ..., m_w I_{m_w})$
\end{definition}
We note that the GIR definition reduces to the Lasso IR condition for singleton groups(see \cite{Basu}).

\begin{definition}[Group beta-min Condition]
 The group beta-min Condition is met for $\beta^0$ , if 
 $\|\beta^{G_r}\|_{\infty} \geq \frac{D \lambda \sqrt{m_r}}{n} \quad \forall r \in W $,
 where $D>0$ is a constant which depends on $\sigma, \phi_{grp}$ and other constants used in cone constraints and GIR conditions. 
\end{definition} 
We note that, only one component of the $\beta^{G_r}, \forall r \in W$ has to be sufficiently large, because we aim to select groups as a whole, and not individual variables. For its exact form, we refer to \cite{Florentina}.

\begin{theorem}
Under the following assumptions the group Lasso selects the true active groups $W$ with high probability:
\begin{itemize}
\item[$(A7)$:] GIR Condition holds for $W$.
\item[$(A8)$:] Group beta-min condition holds for $\beta^{G_r}, \forall r \in W$.
\end{itemize}
\end{theorem}

Next, we discuss sufficient condition for the GIR to hold.
We denote $\Sigma_{r,l} =  X^{G_r^{'}} X^{G_l}/n$,  $r,l \in \{ 1,...,q\}$.
We partition the covariance matrix group wise. (here we assume that each $\Sigma_{r,r}$ is non-singular, or we may use the pseudo inverse)
\begin{align*}
R_W = 
\left[ \begin{array}{cccc}
	I & \Sigma^{-1/2}_{11}\Sigma_{12}\Sigma^{-1/2}_{22} & ... &  \Sigma^{-1/2}_{11}\Sigma_{1w}\Sigma^{-1/2}_{ww}	\\
	\Sigma^{-1/2}_{22}\Sigma_{21}\Sigma^{-1/2}_{11} & I & ...& \Sigma^{-1/2}_{22}\Sigma_{2w}\Sigma^{-1/2}_{ww} \\ 
	\vdots & \vdots & \ddots & \vdots\\
	\Sigma^{-1/2}_{ww}\Sigma_{w1}\Sigma^{-1/2}_{11} & \Sigma^{-1/2}_{ww}\Sigma_{w2}\Sigma^{-1/2}_{22} &... & I 
	\end{array} \right]
\end{align*}
We note that diagonal elements are $I_{m_r \times m_r}$ identity matrix due to parameterization invariance properties.

Now Suppose that $R_W$, has smallest eigenvalue $\Lambda_{min}(R_{W}) > 0$ and that canonical correlations between groups are small enough that intern implies the incoherence assumptions. Therefore, under the eigenvalue and incoherence condition of $R_W$, the group irrepresentable condition holds(see \cite{Buhlmann2}).

Now we prove that the Lasso IR condition implies the group Lasso IR(GIR) condition.

\begin{lemma}
	Let $W \subset \{ 1,...,q \}$ be a group index set, say, $W = \{ 1, ..., w\}$ Consider the full index set corresponding to W:
	 \[ S = \{ (1,1), ..., (1,m1),..., (w,1),...,(w,m_w) \} = \{ 1,...,s \}\] where $s = \sum^{w}_{j=1} m_j$. If the Lasso IR condition holds for the set S then the group Lasso IR condition holds for the set W.
\end{lemma}
Proof is trivial, the IR condition on the set S implies that $ \Sigma_{11}$ is invertible , $\Lambda_{min}(\Sigma_{11})>0$, and correlation between variables in S and between variables in $S$ and $S^c$ are small enough. That implies small enough canonical correlations within the groups in active groups W, and between the groups in $W$ and $W^c$. The small enough canonical correlations between groups ensure the incoherence assumptions and therefore the GIR condition holds.

The following inequality shows the similar bounds for prediction error and estimation error of the group Lasso estimator(\cite{Buhlmann2}).
 \begin{align*}
 	 \frac{1}{n}\| X \Delta \|_{2}^{2} +  \lambda \sum_{r =1}^{q}\| \Delta_{G_r} \|_{2}  & \leq \frac{24\lambda^2 \sum_{r \in W} m_r}{ \phi^{2}_{grp}(W)}
 \end{align*}

\section{The Adaptive Cluster Lasso Methods}
It is known that the Lasso tends to select one or few variables from the group of highly correlated variables, even though many or all of them belong to the active set. We aim to avoid false negatives and solve clustering problem efficiently and more accurately. To solve clustering problem efficiently, we propose a preprocessing step to reduce the dimensionality using the Lasso methods before clustering of variables. To avoid the false negative, we use the concept of clustering the correlated variables and then selecting or dropping the whole group instead of single variables same as the CGLcor(CRLcor) proposed in \cite{Buhlmann2}. 
 The proposed procedure, ACL is a 3-stage procedure, where we can choose to use different methods at different stages depending on the nature of the problem.
The different stages of the ACL procedure is explained as follows.

\begin{enumerate}
\item Dimensionality Reduction \\
For selecting initial set of variables, we use the Lasso(or its adaptive or thresholded version). Since the Lasso tends to select one or a few variables from the group of strongly correlated variables, 
therefore we use the Lasso to select the group representative predictors.  
After we have selected the initial set of variables(group representative members), we get the rest of the group members by simple correlated screening. In section 3, we have shown that for highly correlated structures, the variables set selected by this approach always contains the true active set under assumption of GIR and GBM on the design matrix. 
Let the variables set selected by Lasso is given by
\begin{align}
	\hat{S}_{Lasso} &= \{j; \hat{\beta}_{Lasso,j}(\lambda_1) \neq 0\} .
\end{align}
Then we select correlated variables as
\begin{align*}
	\hat{S}_{corr} &= \{k; \quad k \in \{1,...,p \}\setminus \hat{S}_{Lasso}, j \in \hat{S}_{Lasso} \text{ and }corr(X_j, X_k) > \rho\},
\end{align*}

where $\lambda_1$ is the tuning parameter used by Lasso and $\rho > 0.7$ denotes the strong correlation between two variables.
Then the selected set of variable are given by
 \begin{align}
	\hat{S}_{1} &=  \hat{S}_{Lasso} \cup \hat{S}_{corr}
\end{align}
\item Clustering of Variables\\
After first stage there may be huge amount of reduction in the dimensionality, we denote the reduced design matrix as \textbf{$X_{red} =  \{X_j; j \in \hat{S}_{1}\}$}. On the reduced set of predictors, we apply correlation based clustering methods to group strongly correlated variables into disjoint groups. 
We denote the inferred clusters as $G_1, . . . , G_q $.

\item Supervised Selection of Clusters\\
From the reduced design matrix \textbf{$X_{red}$}, and inferred clusters $G_1, . . . , G_{q}$ as described in previous stages, we select the variables in a group-wise fashion which involves selecting or dropping the group as a whole.  Various methods have been proposed to achieve grouping effect in case of highly correlated variables, i. e. the group Lasso(\cite{Yuan}), Group Square-Root Lasso(\cite{Florentina}), Adaptive group Lasso(\cite{Wei}) and Lasso on cluster representatives etc. 


Suppose, the selected set of groups are denoted by  
\[\hat{S}_{G} = \{r: \text{ group } G_r \text{ is selected}\}\]
The final selected set of variables is then the union of the selected groups.
\[
	\hat{S}_{ACL} = \cup_{r} r \in \hat{S}_{G} 
\]

\end{enumerate}

\begin{algorithm}[h!]
\SetAlgoLined
\textbf{Input:} dataset $(Y,X)$\\
\textbf{Output:} $\hat{S}$:= set of selected variables\\
\caption{ACGL Algorithm}\label{fwd}
\textbf{Steps:}
  Perform Lasso on data $(Y,X)$, Denote $\hat{S}_{Lasso}$ as variable set selected \\
   $S_1$ := $\hat{S}_{Lasso}$ \\
  \For { $ \textbf{ each } j \in \hat{S}_{Lasso}$}{
 	 \For { $ \textbf{ each } k \in \{ 1,...,p\}  /\hat{S}_{Lasso}$}{
  
 		 \If{ $ corr(X_j,X_k) > \rho$ }{
    		$ S_1 = S_1 \cup \{ k \}$ 
    		} 
  
  	  } 
	}
    Let $X_{red} = X^{S_1}$ be the reduced design matrix \\
		
		Perform Clustering of variables on data $X_{red}$, \\
		Denote clusters as $G_1, ..., G_q$  and partition variable set as\\ 	
		$\hat{S_{G_1}}$, ..., $\hat{S_{G_q}}$\\
		Perform group Lasso on $(Y,X_{red})$ with group information $G_1, ..., G_q$, denote the selected set of groups as \\
		$\hat{S}_{cluster} = \{r; \text{ cluster } G_r \text{is selected, } r = 1, . . . , q\}.$\\
		 The union of the selected groups is then the selected set of variables\\
		 $\hat{S}_{ACL} = \cup_{r} r \in \hat{S}_{cluster} $\\
  \textbf{return} $\hat{S}$ 
\end{algorithm}

\subsection{Complexity Analysis of the ACL Method}
In this section, we compute time complexity of the ACL method at different stages.

\begin{enumerate}
\item First stage \\
The time complexity of the first stage consists of the time complexity of the Lasso plus time required for variable screening. 
suppose $\hat{s} = |\hat{S}_{Lasso}|$, denotes the number of variables selected by Lasso, then time taken by variable screening is $\hat{s}*(p-\hat{s})$, which is $O(p\hat{s})$. \\
\item Second stage\\
 Computationally, The second step can be completely avoided. Clustering can be done while screening correlated variables at the first stage itself. However, We opted to state the clustering method separately for transparency and for deriving its theoretical properties, in particular comparing it with other methods where clustering is performed at different stages i.e. CGLcor. But while implementing the algorithm, we can efficiently combine variable screening and clustering. So no extra computational cost is added at this stage.\\
\item Third stage\\
The same computational complexity as for the group Lasso, which depends on the number of groups and size of each group.
\end{enumerate}

Hence the overall time complexity of the proposed method is dominated by the time required for the Lasso and the group Lasso, see \cite{Julien} for complexity analysis of the Lasso.

\section{Theoretical Properties of the ACL Procedure}

In this section, we study the theoretical properties of the ACL methods, and we show that nothing is lost by using ACL methods instead of Clustering Lasso methods with correlation based clustering as proposed in \cite{Buhlmann2}. Particularly, under the GIR and group beta-min condition, the ACGL method has the same accuracy as the CGLcor, in terms of estimation, prediction and variable selection. The gain is in terms of computations since the ACGL performs clustering on the reduced set of predictors.




We introduce the following theorems which are needed for proving variable selection consistency for the proposed algorithm.

\begin{theorem}[Reduced-set IR condition for the Lasso]
Suppose that, the uniform-$\theta$ IR condition is met for the true active set $S := S_0$, which in turn implies that with large probability, the Lasso does not make false positive selection of variables. Then for any $S_1 \subset S $, the uniform-$\theta_1( \leq \theta)$ IR condition holds for the set $S_1$.
\end{theorem}
\textbf{Proof}
We invoke the result given in the book \cite{Buhlmann1}, Corollary 7.2.
Since the uniform-$\theta$ IR condition holds for the set S, then the following inequality also holds.
\[
	\frac{\sqrt{s} \mathop{max}_{j \not\in S} \sqrt{\sum_{k \in S} \sigma^{2}_{jk}}} {\Lambda^{2}_{min}(\Sigma_{11}(S)) } \leq \theta
\]
where $\sigma_{jk}$ denotes the $(jk)^{th}$ entry of $\Sigma$.\\
Now we delete some variables from S, and denote the reduced subset as $S_1$ and corresponding partition of variance-covariance matrix is $\Sigma_{11}(S_1)$ and $\Sigma_{21}(S_1)$. Since $S1 \subset S$, the following inequality holds because any symmetric minor of $\Sigma_{11}(S)$ will have min-eigenvalues at least as big as  $\Lambda_{min}(\Sigma_{11}(S))$.
\[
\Lambda_{min}(\Sigma_{11}(S_1)) \geq \Lambda_{min}(\Sigma_{11}(S)) 
\]
Therefore 
\[
	\theta_1 = \frac{\sqrt{s_1} \; max_{j \not\in S_1} \sqrt{\sum_{k \in S_1} \sigma^{2}_{jk}}} {\Lambda^{2}_{min}(\Sigma_{11}(S_1)) } \leq
	\frac{\sqrt{s} \; max_{j \not\in S} \sqrt{\sum_{k \in S} \sigma^{2}_{jk}}} {\Lambda^{2}_{min}(\Sigma_{11}(S)) } \leq \theta
\]
Hence the IR condition holds for the set $S_1 \subset S$.

The similar result holds for the group Lasso which is given in the following lemma.
\begin{lemma}[Reduced-group IR condition for the group Lasso]
Suppose that, the uniform-$\theta$ IR condition is met for the set of true active groups $W \subset \{1,...,q \}$, It implies that with large probability, the Lasso does not make false positive selection of groups. Then the uniform-$\theta_1( \leq \theta)$ GIR condition holds for the following cases:
\begin{itemize}
	\item for any $W_1 \subset W $, when the number of groups are reduced.
	\item when $|W_1| = |W| $ but $\{G_r, r \in W_1\} \subset \{ G_r, r \in W\} $, group sizes are reduced for some groups.
	\item when group size as well as number of groups are reduced, $W_1 \subset W $ and $\{G_r: r \in W_1\} \subset \{ G_r: r \in W \} $.
\end{itemize}
\end{lemma}

Proof is trivial. Suppose that IR condition holds for the active set $S$. If w disjoint groups are formed within S such that $ S= \sum_{j \in W} m_j$, then IR for the Lasso imply IR condition for the group Lasso. Since the Reduced-set IR will hold for $S_1 \subset S$, where the reduced set can be interpreted as change in group structure in terms of reduced number of groups and/or reduced size of groups, as

\begin{align*}
	S_1 &= \sum_{r \in W_1} m_r, \quad W_1 \subset W 
\end{align*}
Therefore Reduced-group IR will hold for the set $S_1$.

\subsection{Case Studies}
In this section, we illustrate the variable selection consistency of the  proposed method using a couple of scenarios under assumption of GIR, group beta-min and no noise case. 
\subsubsection{Orthonormal Case} 
The case $ \Sigma \approx I $ corresponds to uncorrelated variables and hence IR condition holds for any $S \subset \{ 1,...,p\}$ and we also assume that beta-min condition holds for a fixed S. 

We claim that ACGL selects the true active set $\hat{S}_{ACGL} = S_0$ with high probability.\\
Proof:\\
First, we perform the Lasso operation on the pair (Y,X), to get the initial set of variables, say $\hat{S}_1$. Then with high probability $\hat{S_1} =  S_0$ under assumption of IR and beta-min condition. Since variables are uncorrelated no additional variable will be pulled in when we do correlation screening. At the second stage, we perform clustering on the reduced set of predictors. Clustering process will report the singleton groups due to independence structure, and finally at the third step CGL will select the true active set $S_0$ again, due to reduced-set IR condition.

We proved that ACGL consistently selects the true active set for the orthonormal case. It is obvious that, there is no advantage of using AGCL or the plain CGL/CRL methods over the standard Lasso for this case. But ACL outperforms over CGL/CRL in terms of computations, since AGCL considers the reduced set of predictors for clustering and the group lasso is called for reduced number of groups, whereas plain CGL/CRL considers all p variables for clustering which requires huge computations when the dimension is ultra high.
\subsubsection{Block Diagonal Case}
The case $ \Sigma \approx diag(\mathcal{T}_1, \mathcal{T}_2, ..., \mathcal{T}_q) $ corresponds to uncorrelated groups and hence group IR condition holds for any $W \subset \{ 1,...,q\}$ and we also assume that group beta-min condition holds. Each $\mathcal{T}_i$ is a ${m_i \times m_i}$ matrix with elements as 
\[
	(\mathcal{T}_i){j,k} =  \left\lbrace \begin{array}{cc}
	1, & j=k\\
	\rho, & else	
	\end{array} \right.
\] where $\rho > 0.7$, since variables are highly correlated within each group.
Without loss of generality we can assume that all the variables are ordered in a way such that all active groups come first.  $W \subset \{1,...,q\} $ then $W = \{1,...,w \}. $ and we also assume pure active or pure noise group. 
Let
\begin{align*}
	S_0 &= \{ (1,1),..., (1,m_1), ..., (w,1),..., (w,m_w) \} = \{1,...,s_0 \},
\end{align*}
be the true active set.
The Lasso tends to select(depending on the amount of regularization) one variable from each active block. 
Without loss of generality we assume that the Lasso selects (j,1) variable from each $j \in W$, and the selected variable set is $\hat{S}_{Lasso} = \{ (1,1),(2,1),...,(w,1) \}$,  
	Now we add all variables from $\{1,...,p \}/S_{Lasso}$ which are strongly correlated with atleast of the the variable from $S_{Lasso}$. Therefore we get 
	\begin{align*}
		\hat{S}_1 &= \{1,...,s \}\\
		\implies & \hat{S}_1 = S_0
	\end{align*}
	Hence, after the first stage of dimensionality reduction, the selected set of variables contains the true active variables.
	Assuming that the clustering procedure correctly identifies the true underlying group structure, then the group Lasso at the third stage correctly selects all the w groups, due to the sub-group IR condition for the group Lasso. 
	Hence, the proposed method consistently selects the true active set under the assumption of GIR and group beta-min condition for the block diagonal case as well.

One may argue that, there is no need for the second and the third stage. Specifically, when the Lasso selects one variable per active group then correlation screening will bring in those correlated variables which were not selected by the Lasso. We refer to the discussion in \cite{Junzhou}, on using the group Lasso over Lasso.

\section{Numerical Results}
In this section, we consider three simulation settings and a pseudo real data example in order to empirically compare the performance of the proposed method with the other existing methods. 
Since the comparison between the Lasso, CGL and CRL have already been studied in the paper \cite{Buhlmann2}, here we only report the results for ACGL and CGLcor methods.

\subsection{Simulation Study}
Three examples are considered in this simulation. In each example, data is simulated from the linear model in (equation 1) with fixed design \textbf{X}. All the three examples are the same as simulation examples used in the paper \cite{Buhlmann2}. 

For each example, our simulated dataset consisted of a training and an independent validation set and 50 such datasets were simulated for each example. The models were fitted on the training data for each 50 datasets and the model with the lowest test set Mean Squared Error(MSPE) was selected as the final model. For model interpretation, we consider true positive rate as a measure of performance. We also measure the CPU time expended by each methods. The MSE and the true positive rate are defined as follows.
\begin{align}
    MSE &= \frac{1}{n} \sum_{i=1}^{n} (y_i - \hat{y}_i)^2\\
	TPR &= f(|\hat{S}|) = \frac{ | \hat{S} \bigcap S_0  |} {|S_0|}
\end{align}

\subsubsection{Block Diagonal Model}
\begin{figure}[h!] 
	\includegraphics[scale=.75]{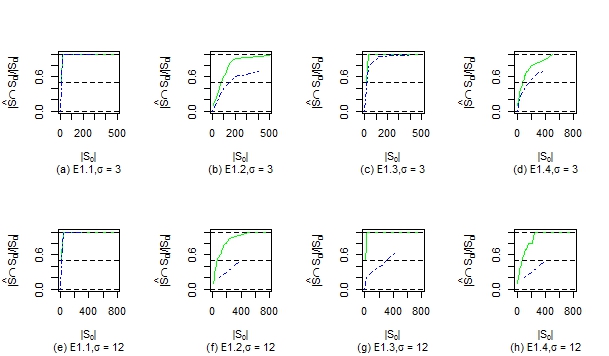}
	\caption{Plot of $ \frac{ | \hat{S} \bigcap S_0  |} {|S_0|}$ vs. $|\hat{S}|$ for block diagonal model. ACGL(green solid line) and CGLcor(blue dashed-dotted line)} \label{fig:E1}
\end{figure}

We generate covariates from $N_p(0,\Sigma_1)$, where $\Sigma_1$ consists of 100 block matrices $\mathcal{T}$, and $\mathcal{T} $ is a ${10 \times 10}$ matrix, defined as
\[
	\mathcal{T}_{j,k} =  \left\lbrace \begin{array}{cc}
	1, & j=k\\
	.9, & else	
	\end{array} \right.
\]
For the regression coefficient $\beta^0$ the following four configurations are considered:\\
(E1.1)
$S_0 = \{1, 2, . . . , 20\}$ and for any $j \in S_0$ we sample $\beta^0_{j}$ from the set $\{.1,.2,.3, . . . , 2\}$ without replacement (a new for each simulation run). This set up has all the active variables in the first two blocks of highly correlated variables.\\
(E1.2) 
$S_0 = \{1,2,11,12 . . . , 91,92\}$ and for any $j \in S_0$ we sample $\beta^0_{j}$ from the set $\{.1,.2,.3, . . . , 2\}$ without replacement (a new for each simulation run). This set up has all the active variables in the first and the second variables of the first ten blocks.\\ 
(E1.3)  The $\beta^0$ has the same configuration as in (E1.1) but we change the sign of randomly chosen half of the active parameters (a new for each simulation run).\\
(E1.4) The $\beta^0$ has the same configuration as in (E1.2) but we change the sign of randomly chosen half of the active parameters (a new for each simulation run).\\
Simulation results are reported in table \ref{table:E11}(MSE and standard deviation), figure \ref{fig:E1}(TPR) and table \ref{table:E13}(CPU time). 
\begin{table}[h!] \centering
		\begin{tabular}{| c c c c c c |}
		\hline 
		$\sigma$ & Method &  E1.1 & E1.2 & E1.3 & E1.4 \\
		\hline 
		3& ACGL & 12.46 (1.76) & 21.95 (2.98) & 9.308 (2.40) & 20.90 (4.81) \\		
		& CGLcor & 14.97 (2.40)& 37.05 (5.21)& 13.34 (2.06)& 24.31 (6.50)\\
		\hline
		12& ACGL & 188.23 (23.32) & 149.97 (28.98) & 129.29 (22.93) & 165.91 (22.36)\\
		& CGLcor & 206.19 (29.97) &186.61 (25.69) &160.31 (23.04) &168.26 (24.70)\\
		\hline 	
		\end{tabular}	
		\caption{MSE(sd) for Example block diagonal model}\label{table:E11}
\end{table}

\begin{table}[h!] \centering
		\begin{tabular}{| c c c c c c|}
		\hline 
		$\sigma$ & Method & E2.1 & E2.2 & E2.2 & E2.4  \\
		\hline 
		3& ACGL &4.25  &  5.04  & 2.09 & 6.46 \\	
		& CGLcor & 2510 & 2510 & 2510 & 2510 \\
		\hline
		12 & ACGL & 3.91 & 5.86  & 1.08 & 3.33 \\
		& CGLcor & 2510 & 2510 & 2510 & 2510 \\
		\hline 	
		\end{tabular}	
		\caption{CPU times(in seconds) for block diagonal model}\label{table:E13}
\end{table}
From table \ref{table:E11}, we see that the ACGL method has lower prediction error than the CGLcor for all four configurations and figure \ref{fig:E1} shows that the ACGL has higher TPR. From table \ref{table:E13} we see that ACGL is much efficient, the CPU time required for ACGL for all four configurations are much less than as compared to the CPU time expended by CGLcor. Please note that CPU time for CGLcor is approximately the same for all configurations.

\subsubsection{Single Block Design}
We generate covariates from $N_p(0,\Sigma_2)$, where $\Sigma_2$ consisted of a single group of strongly correlated variables of size 30, it is defined as
\[
	\Sigma_{2;j,k} =  \left\lbrace \begin{array}{cc}
	1, & j=k\\
	 0.9& i, j \in \{1, . . . , 30\} \text{ and } i != j,	\\
	 0& otherwise
	\end{array} \right.
\]
The remaining 970 variables are uncorrelated. For the regression coefficient $\beta^0$ we consider the following four configurations:\\
(E2.1) $S_0 = \{1, 2, . . . , 15\} \cup \{31, 32, . . . , 35\}$ and for any $j \in S_0 $ we sample $\beta^0_{j}$ from the set $\{.1,.2,.3, . . . , 2\}$ without replacement (new for each simulation run). The correlated block contains 15, the most of the active predictors and the remaining five active predictors are uncorrelated. \\
(E2.2) $S_0 = \{1, 2, . . . , 5\} \cup \{31, 32, . . . , 45\}$ and for any $j \in S_0 $ we sample $\beta^0_{j}$ from the set $\{.1,.2,.3, . . . , 2\}$ without replacement (new for each simulation run). Here the correlated block contains only 5 active predictors, and the remaining 15 predictors are uncorrelated.\\
(E2.3) The $\beta^0$ has the same configuration as in (E2.1) but we change the sign of randomly chosen half of the active parameters (new for each simulation run).\\
(E2.4) The $\beta^0$ has the same configuration as in (E2.2) but we change the sign of randomly chosen half of the active parameters (new for each simulation run).

\begin{figure}[h!] 
	\includegraphics[scale=.75]{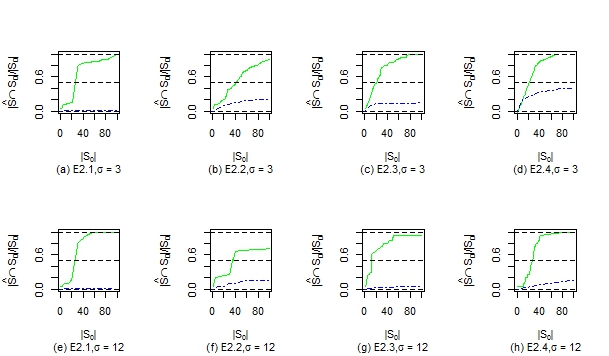}
	\caption{Plot of $ \frac{ | \hat{S} \bigcap S_0  |} {|S_0|}$ vs. $|\hat{S}|$ for single block model. ACGL(green solid line) and CGLcor(blue dashed-dotted line)} \label{fig:E2}
\end{figure}

Simulation results are reported in table \ref{table:E21}, table \ref{table:E23} and figure \ref{fig:E2}.
\begin{table}[h!] \centering
		\begin{tabular}{| c c c c c c|}
		\hline 
		$\sigma$ & Method & E2.1 & E2.2 & E2.3 & E2.4  \\
		\hline 
		3& ACGL & 11.40 (4.2) & 29.94 (5.34)&  15.01 3.28) & 27.03 (3.9)\\		
		& CGLcor & 247.52 (28.74) & 54.73 (10.59) & 21.37 (9.51) &31.58 (14.17)\\
		\hline
		12& ACGL & 146.17(23.46) & 192.64 (12.81)& 127.91 (22.02)& 159.62 (26.40)\\
		& CGLcor & 384.78 (48.26) &191.26 (25.55) &159.40 (23.88) &174.49 (25.40)\\
		\hline 	
		\end{tabular}	
		\caption{MSE(sd) for single block model}\label{table:E21}
\end{table}

\begin{table}[h!] \centering
		\begin{tabular}{| c c c c c c|}
		\hline 
		$\sigma$ & Method & E2.1 & E2.2 & E2.3 & E2.4  \\
		\hline 
		3& ACGL & 7.58 & 5.52 & 2.63   & 5.63  \\	
		& CGLcor & 2463 & 2463 & 2463 & 2463 \\
		\hline
		12& ACGL & 3.17  & 4.92  & 2.37&  2.47  \\
		& CGLcor & 2463 & 2463 & 2463 & 2463 \\
		\hline 	
		\end{tabular}	
		\caption{CPU times(in seconds) for single block model}\label{table:E23}
\end{table}

Table \ref{table:E21} shows that the ACGL method has lower predictive performance than the CGLcor for all four configurations and figure \ref{fig:E2} shows that in terms of variable selection, ACGL is clearly better than CGLcor. From table \ref{table:E21}, we see that CPU time required(in seconds) for ACGL for all four configurations are much less than as compared to the CGLcor. The CPU time for CGLcor is approximately the same for all configurations.
 
\subsubsection{Duo Block Model}
\begin{figure}[h!]
\begin{center}
	\includegraphics[scale=.75]{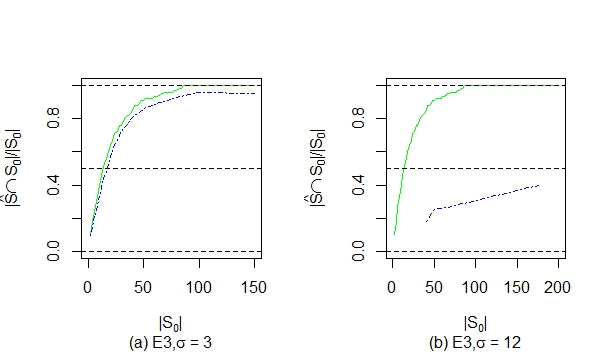}
\end{center}
	\caption{Plot of $ \frac{ | \hat{S} \bigcap S_0  |} {|S_0|}$ vs. $|\hat{S}|$ for duo block model. ACGL(green solid line) and CGLcor(blue dashed-dotted line)}  \label{fig:E3}
\end{figure}

We generate covariates from $N_p(0,\Sigma_3)$, where $\Sigma_3$ consists of 500 block matrices $\mathcal{T}, $ and $\mathcal{T} $ is a ${2 \times 2}$ matrix, defined as
\[
	\mathcal{T}_{j,k} =  \left\lbrace \begin{array}{cc}
	1, & j=k\\
	.9, & else	
	\end{array} \right.
\]
For the regression coefficient $\beta_0$ we consider $S_0 = \{1, 2, . . . , 20\}$ and for any $j \in S_0$ 
\[
	\beta^0_{j} =  \left\lbrace \begin{array}{cc}
	2, &  j \in \{1, 3, 5, 7, 9, 11, 13, 15, 17, 19\},\\
	\frac{\frac{1}{3} \sqrt{\frac{\log p}{n}} \sigma} {1.9}, &  j \in \{2, 4, 6, 8, 10, 12, 14, 16, 18, 20\}	
	\end{array} \right.
\]
In this setup, the $\beta^0$ is the same for all 50 simulation runs. The Lasso would not select the variables from $\{2, 4, 6, . . . , 20\}$, since they do not satisfy the beta-min condition but it would select the other from $\{1, 3, 5, . . . , 19\}$. The Table \ref{table:E31}, Table \ref{table:E32} and Figure \ref{fig:E3} show the simulation results
for the duo block model.

\begin{table}[h!] \centering
		\begin{tabular}{| c c c |}
		\hline 
		$\sigma$ & Method & MSE(sd)  \\
		\hline 
		3& ACGL & 20.82 (5.94)\\	
		& CGLcor & 32.00 (6.50) \\
		\hline
		12& ACGL & 179.11 (19.35)\\
		& CGLcor & 193.97 (27.05)  \\
		\hline 	
		\end{tabular}	
		\caption{MSE(sd) for duo block model}\label{table:E31}
\end{table}

\begin{table}[h!] \centering
		\begin{tabular}{| c c c |}
		\hline 
		$\sigma$ & Method & CPU time  \\
		\hline 
		3& ACGL & 7.69\\	
		& CGLcor & $>$ 2500 \\
		\hline
		12& ACGL & 2.94\\
		& CGLcor & $>$ 2500  \\
		\hline 	
		\end{tabular}	
		\caption{CPU time(in seconds) for duo block model}\label{table:E32}
\end{table}
The results show that the ACGL performs better in terms of predictive performance, variable selection and CPU time required. (We stopped recording CPU time for CGLcor after 2500sec, here clustering of a thousand of variables and then the group lasso for $\approx$500 clusters make the process very slow.
\subsection{Pseudo Real Data}
We consider here a real dataset, riboflavin$(n = 71, p = 4088)$ data for the design matrix X with synthetic regression coefficients $\beta^0$ and simulated Gaussian errors $N_n(0, \sigma^2 I)$. See \cite{HDview} for the details on riboflavin dataset. 
We select the first thousand covariates which have largest empirical variances. We fix the size of the active set to $s_0 =10$. 
For the true active set, we randomly select a variable, say variable k(a new in each simulation), and then we select other nine variables which have highest absolute correlation to the variable k, and for each $j \in S_0$ we set $\beta_j = 1$. This configuration is exactly the same as pseudo real example used in \citep{Buhlmann2}.

\begin{figure} [h!]
\begin{center}
	\includegraphics[scale=.75]{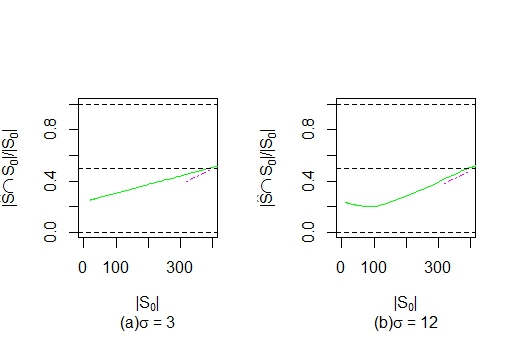}
\end{center}
	\begin{flushleft}
	\caption{Plot of $ \frac{ | \hat{S} \bigcap S_0  |} {|S_0|}$ vs. $|\hat{S}|$ for Riboflavin dataset. ACRL(green solid line) and CRLcor(magenta dashed-dotted line)}\label{fig:Ribo}
	\end{flushleft}
\end{figure}

The performance measures are reported in table \ref{table:Ribo1}, figure \ref{fig:Ribo} and table \ref{table:Ribo2}, based on 50 independent simulation runs. Here we compare CRLcor(\citep{Buhlmann2}) with Adaptive Cluster Representative Lasso(ACRL) where the Lasso is used as preprocessing step at the first stage, correlation based clustering at second step and the Lasso for cluster representatives in the third stage. The Group Lasso is not appropriate for this setup, since k is chosen arbitrarily and the group size may exceed the number of observations.

\begin{table}[h!] \centering
		\begin{tabular}{| c c c |}
		\hline 
		$\sigma$ & Method& MSPE(std) \\
		\hline 
		3& ACRL & 2.36 (0.52) \\	
		& CRLcor & 39.02 (25.15) \\	
		\hline
		15& ACRL & 25.02 (4.03)   \\
		& CRLcor & 50.40 (27.68) \\
		\hline
		\end{tabular}	
		\caption{MSE(sd) for Riboflavin dataset}\label{table:Ribo1}
\end{table}

\begin{table}[h!] \centering
		\begin{tabular}{| c c c |}
		\hline 
		$\sigma$ & Method & E3  \\
		\hline 
		3& ACGL &  1.47\\	
		& CRLcor & 2347 \\
		\hline
		12& ACGL & 1.48 \\
		& CRLcor & 2471  \\
		\hline 	
		\end{tabular}	
		\caption{CPU times(in seconds) for Riboflavin dataset}\label{table:Ribo2}
\end{table}

The table \ref{table:Ribo1}, figure \ref{fig:Ribo} and table \ref{table:Ribo2} show that ACRL performs better than CRLcor in terms of prediction, variable selection and CPU time consumption.
 
\section{Computational Details}
Statistical analysis was performed in R 3.2.2. We used, the package ``glmnet" for penalized regression methods(the Lasso and adaptive Lasso), the package ``gglasso" to perform group Lasso and the package ``ClustOfVar" for clustering of variables.
\section{Conclusion and Future Work}
In this article, we proposed a three stage procedure 
for variable selection for high-dimensional linear model with strongly correlated variables. Our procedure is an extension of the algorithms proposed in \cite{Buhlmann2}. A technical extension compared with \cite{Buhlmann2} is that we propose to reduce the dimension at the first stage using Lasso(or its adaptive or thresholded version) prior to clustering at the second stage and then supervised selection of clusters in the third stage. 
 We proved that the variables selected by our algorithm contains the true active set consistently under GIR and group beta-min conditions. Our simulation studies show that reducing dimension improves the speed and accuracy of the clustering process and then considering correlation structure improves variable selection and predictive performance.

Since the theoretical results we developed for our algorithms are not restricted to the squared error loss, it can be extended to the generalized linear models, i.e, the preprocessing step of dimensionality reduction can be added to the group Lasso for the logistic regression(\cite{Meier}), and this is our future work.

\section*{References}
  \bibliographystyle{elsarticle-harv} 
  \bibliography{niharika_arXiv}

\end{document}